\documentclass[a4paper,fleqn]{cas-sc}
\usepackage{fancyhdr}
\usepackage{geometry}

\usepackage{graphicx} 
\usepackage[authoryear,longnamesfirst]{natbib}
\usepackage{hyperref}
\usepackage{subcaption}  
\usepackage{amsmath}
\usepackage{amssymb}
\usepackage{comment}
\usepackage{multirow} 
\usepackage{booktabs}
\usepackage{threeparttable} 
\usepackage{lineno}
\usepackage{lscape}
\usepackage{setspace}
\usepackage[dvipsnames]{xcolor}
\usepackage{caption}

\def\bX{\mathbf{X}}

\def\bSigma{{\boldsymbol\Sigma}}

\def\bz{\mathbf{z}}

\def\bmu{{\boldsymbol\mu}}

\def\bphi{{\boldsymbol\phi}}

\def\bpsi{{\boldsymbol\psi}}

\def\cN{\mathcal{N}}

\def\exp{{\sf exp}}

\usepackage{etoolbox}
\makeatletter

\patchcmd{\maketitle}{\thispagestyle{first}}{\thispagestyle{fancy}}{}{}

\def\ps@first{%
  \def\@oddfoot{}%
  \def\@evenfoot{}%
  \def\@oddhead{}%
  \def\@evenhead{}%
}

\makeatother

\begin{document}
\let\WriteBookmarks\relax
\def\floatpagepagefraction{1}
\def\textpagefraction{.001}

\shorttitle{}    

\shortauthors{}  

\title [mode = title]{Fairness in Machine Learning-Based Hand Load Estimation: \\A Case Study on Load Carriage Tasks}

%

\author[1]{Arafat Rahman}
\address[1]{Department of Systems and Information Engineering, University of Virginia, 151 Engineer’s Way, Charlottesville, VA, USA}

\author[2]{Sol Lim}
\address[2]{Department of Industrial and Systems Engineering, Virginia Polytechnic Institute and State University, 1145 Perry Street, Blacksburg, VA, USA}

\author[1]{Seokhyun Chung}[orcid=0000-0001-5176-4180] 
\ead{schung@virginia.edu}
\cormark[1]
\cortext[1]{Corresponding author}


\begin{abstract}
Predicting external hand load from sensor data is essential for ergonomic exposure assessments, as obtaining this information typically requires direct observation or supplementary data. While machine learning methods have been used to estimate external hand load from worker postures or force exertion data, our findings reveal systematic bias in these predictions due to individual differences such as age and biological sex.
To explore this issue, we examined bias in hand load prediction by varying the sex ratio in the training dataset. 
We found substantial sex disparity in predictive performance, especially when the training dataset is more sex-imbalanced. 
To address this bias, we developed and evaluated a fair predictive model for hand load estimation that leverages a Variational Autoencoder (VAE) with feature disentanglement.
This approach is designed to separate sex-agnostic and sex-specific latent features, minimizing feature overlap. The disentanglement capability enables the model to make predictions based solely on sex-agnostic features of motion patterns,  ensuring fair prediction for both biological sexes.
Our proposed fair algorithm outperformed conventional machine learning methods (e.g., Random Forests) in both fairness and predictive accuracy, achieving a lower mean absolute error (MAE) difference across male and female sets and improved fairness metrics such as statistical parity (SP) and positive and negative residual differences (PRD and NRD), even when trained on imbalanced sex datasets. These findings emphasize the importance of fairness-aware machine learning algorithms to prevent potential disadvantages in workplace health and safety for certain worker populations.
\end{abstract}


\maketitle
\thispagestyle{fancy}

\doublespacing
\section{Introduction}
\label{sec:intro}

Advancements in sensor and monitoring technologies are creating new opportunities to enhance ergonomic risk assessments through data-driven approaches. 
Wearable inertial sensors~\citep{lim2020narrative} and computer vision-based joint tracking systems~\citep{massirisfernandez2020ergonomic} provide real-time, high-resolution monitoring of worker kinematics in workplace settings. 
These technologies capture detailed motion data, facilitating real-time biomechanical assessment~\citep{peppoloni2016novel}, continuous risk monitoring for early intervention~\citep{lorenzini2022online}, and personalized ergonomic recommendations tailored to individual movement patterns and workload conditions~\citep{kim2021directional,lim2023real}.

The availability of such data has spurred efforts to automate musculoskeletal disorder (MSD) risk assessment and mitigation using machine learning (ML). 
By leveraging sensor-derived movement patterns, force exertion data, and workers-specific attributes (e.g., stature and strength), ML models can estimate ergonomic risk factors that are difficult to measure directly. 
For instance, various ML models have been used to estimate key ergonomic risk factors, such as the weight of lifted~\citep{taori2024use,hlucny2020characterizing,lim2024exposures,lim2020classifying} or carried objects~\citep{lim2019statistical,yang2020deep}, as well as the mode of carrying or lifting techniques (e.g., single-hand or two-handed). 
The key premise is that, with sufficient data, ML models can uncover intricate correlations between sensor outputs, worker attributes, and ergonomic risks---potentially augmenting or even replacing traditional assessment methods based on manual observations and standardized checklists.


ML-based MSD risk evaluation has been explored across various occupational domains. In construction,~\citet{antwi2018wearable} achieved 99.70\% accuracy in detecting awkward postures using a wearable insole pressure system and a Support Vector Machine (SVM) model, enabling non-invasive MSD risk monitoring. \citet{mudiyanselage2021automated} used surface EMG sensors and decision tree algorithms to automate ergonomic risk assessments, classifying MSD risks based on the National Institute for Occupational Safety and Health (NIOSH) lifting equation with 99.4\% accuracy. 
For health service workers, \citet{luo2024explainable} introduced an explainable ML model with Boruta feature selection, streamlining neck and shoulder MSD risk screening using 12-17 key items. \citet{trkov2022classifying} combined instrumented insoles and accelerometers to detect material handling activities and assess MSD risks in real-time, achieving 85.3\% accuracy. 
Recent studies have also explored generative models for ergonomic risk assessment. \citet{li2021lifting} developed conditional Variational Autoencoder (VAE) and generative adversarial networks to predict realistic lifting postures from body measurements. \citet{qing2024predicting} introduced U-Net and diffusion models for predicting human lifting postures.  
These approaches highlight the potential of ML in MSD risk assessment.

Despite these successes, biomechanical differences across demographic groups can introduce systematic biases in ML predictions, particularly when certain groups are underrepresented in training data. ML models often fail to generalize well to underrepresented populations due to disparities in training data distribution. Worker demographics (e.g., age, biological sex~\citep{yfantidou2023beyond}) and physical characteristics (e.g., strength) substantially influence movement patterns--for instance, females tend to exhibit greater cadence and shorter stance time than men while carrying loads~\citep{middleton2022mechanical,harper1997female,holewijn1992physiological}.
Real-world training datasets are often skewed, leading to biased model performance. 
If a model is trained predominantly on one demographic group, it may fail to accurately predict movement characteristics of others. 
This issue has been observed in accelerometer-based gait detection models, which performed poorly for older adults when trained primarily on younger individuals~\citep{zhang2019pdmove}.
Similarly,~\citet{lim2019gender} found that ML models trained on inertial measurement unit (IMU) sensor data for external hand load estimation consistently underestimated loads carried by males, revealing sex-based disparities. 
These findings suggest that current data-driven ergonomic assessment systems may inherently reflect demographic biases. 
However, to the best of our knowledge, little to no research has systematically addressed this issue. 
Bridging this gap is essential for improving the generalizability of ML-based risk assessments across diverse worker populations, ultimately enabling fairer and more effective occupational safety interventions.

As an initial step toward developing fair ML algorithms for MSD risk assessment, we investigated potential algorithmic biases in ML models estimating MSD risk factors based on a key worker characteristic--biological sex. Specifically, we aimed to answer the following two research questions in this study: 

    \begin{enumerate}[{(1)}]
        \item\textbf{\textit{Research Question 1}: Do conventional ML algorithms exhibit bias when predicting hand load?}
        
        \textit{Approach}: We quantified bias in ML models predicting carried box weight from gait patterns captured by IMU sensors, varying the sex ratio in the training dataset. We assessed the performance of three conventional ML methods that do not explicitly enhance fairness for underrepresented groups. 

        \item \textbf{\textit{Research Question 2:} If bias is present, can we mitigate it by developing a fair ML algorithm? }
        
        \textit{Approach}: We developed a group-wise fair ML model that accounts for inherent biomechanical differences in kinematics and gait patterns, ensuring equitable performance across demographic groups, even when training data is imbalanced. We then evaluated our model's effectiveness in reducing prediction bias across sex groups using multiple fairness metrics. 

    \end{enumerate}



Ultimately, we aim to foster a fairer and more inclusive use of ML models for ergonomic risk assessments by addressing biases in model predictions based on IMU sensor data.
This will improve both the performance and fairness of ML-based risk assessments, even in the presence of skewed training data.
While fair ML algorithms (e.g., sex bias mitigation) have been actively investigated in the area of facial recognition~\citep{cavazos2020accuracy}, driver injury severity classification~\citep{mafi2018machine}, pedestrian detection~\citep{brandao2019age}, and natural language processing~\citep{sun2019mitigating}, their application to ergonomics and MSD risk assessments remains substantially limited. 
Our work contributes to the broader ergonomic field by highlighting fairness challenges in ML-based human performance and risk assessments, which may systematically under- or over-estimate workers' physical demands and capabilities.

\section{Methods}
\label{sec:methods}


\subsection{Data Description}

We used data previously collected in another study, as reported by~\citet{lim2019combining} and~\citet{lim2020measuring}. The dataset comprises measurements from 22 healthy participants (12 males and 10 females). Participants were aged 18 to 55 years, with an average (SD) age of 33.8 (10.0) years, stature of 1.74 (0.08) m, body mass of 76.1 (13.4) kg, and body mass index (BMI) of 25.1 (3.4) kg/m². 
Participants had no pre-existing back injuries or chronic pain. Each participant provided written informed consent, as approved by the university's institutional review board.

In the main experiment, participants carried a weighted box along a level corridor, covering a 24-meter distance using four common occupational carrying methods. These methods included one-handed carrying with the right and left hand, two-handed side carrying, and two-handed anterior carrying. Each carrying method was tested at three hand load levels: 4.5, 13.6, and 22.7 kg. Participants completed two consecutive trials for each of the twelve loaded conditions (4 carrying methods $\times$ 3 load levels), presented in a randomized order. They were allowed to choose their walking speed to reflect natural adjustments under different load conditions. To minimize fatigue and its potential effects, participants received a two-minute rest break between each walking trial.


Twelve commercial inertial sensors (Biostamp RC, mc10 Inc., Lexington, MA, USA) were placed on participants at specific anatomical locations: the left thigh, right thigh, left shank, right shank, right dorsal foot, left upper arm, right upper arm, left forearm, right forearm, the sixth thoracic vertebra (T6), sternum, and the first sacral vertebra (L5/S1). The sensors on the right and left shanks were used to identify key gait events. The inertial sensors recorded 3-axis acceleration and angular velocity at each anatomical location at a sampling frequency of 80 Hz.

\subsection{Data Preprocessing}
\begin{figure*}
    \centering
    \includegraphics[width=1\textwidth]{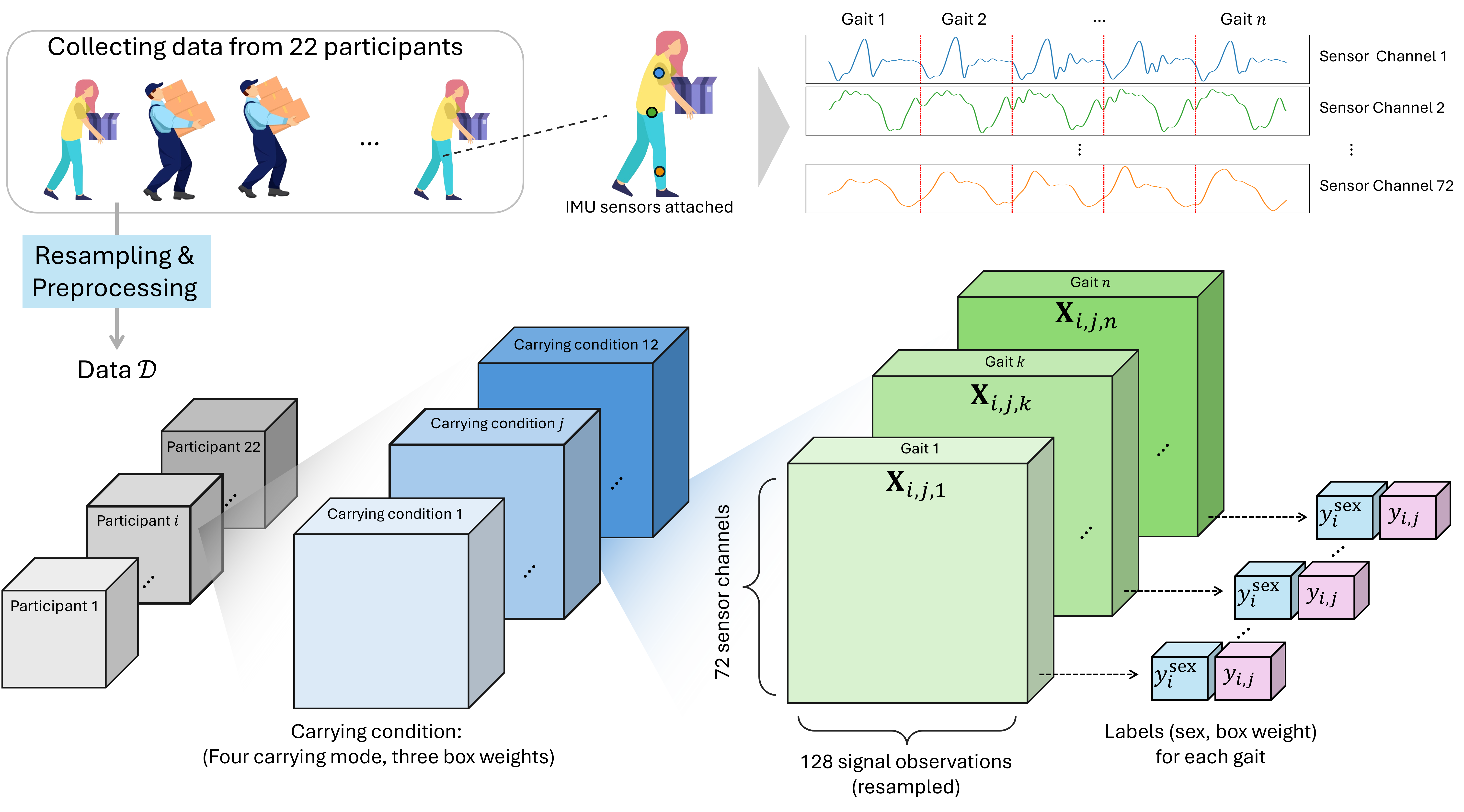}
    \caption{Overview of data structure: Gait pattern data collected from 22 participants across three different box weights (4.5, 13.6, and 22.7 kg), labeled by biological sex and box weight.}
    
    
    \label{fig:data_preprocess}
\end{figure*}

Our data preprocessing steps are illustrated in Figure~\ref{fig:data_preprocess}.
First, we detected gait cycles from continuous inertial sensor data~\citep[for more details, see][]{lim2019statistical}. Specifically, we identified key gait events---heel strikes and toe-offs---using angular velocity data (rad/s) recorded from sensors placed on both the right and left shanks. 
Each gait cycle was defined as a sequence of events: right heel strike → left toe-off → left heel strike → right toe-off → next right heel strike. 
All inertial sensor data were filtered using a second-order low-pass zero-lag Butterworth filter with a 6-Hz cut-off frequency. 
Since gait cycle duration varied, we resampled all cycles to a uniform length of 128 signal samples using 1D linear interpolation. 
Each inertial sensor provided six channels of data (three-axis linear acceleration and three-axis angular velocity), and with 12 sensors in total, this resulted in 72 channels. 
Overall, we collected 4,046 gait cycles across all participants and carrying conditions.
The data were structured into a $4046 \times 128 \times 72$ matrix, where $4046$ represents the total number of gait cycles for all subjects, $128$ is the standardized signal length, and $72$ is the total number of sensor channels.
Box weights (4.5, 13.6, and 22.7 kg) served as the output labels for each gait cycle.  
Although participants used four different carrying modes, we did not include them as output labels. 
Instead, we aggregated all carrying conditions and focused solely on predicting the box weights. 

\subsection{Machine Learning Models}
To examine whether conventional ML algorithms exhibit bias (\textbf{\textit{Research Question 1}}), we compared three commonly used ML methods for hand load estimation. 
We chose $k$-nearest neighbors ($k$-NN), support vector machine (SVM), and Random Forest (RF) due to their diverse learning strategies and established effectiveness in sensor-based prediction tasks~\citep{ye2024machine}. $k$-NN is a non-parametric method that classifies data points based on the majority vote of their nearest neighbors, making it well-suited for capturing local structures in high-dimensional spaces of IMU data~\citep{mohsen2021human}. SVM employs hyperplane-based separation, leveraging kernel functions to model complex relationships in the IMU data~\citep{hearst1998support}. RF, an ensemble learning approach, constructs multiple decision trees to enhance predictive accuracy and reduce overfitting~\citep{breiman2001random}. It is a representative of classical yet popular ML approaches, commonly employed in ML-based ergonomics risk evaluation studies~\citep{aliabadi2022investigation, lim2019statistical, mudiyanselage2021automated}. These models do not explicitly incorporate fairness enhancements for underrepresented groups and were therefore used as baseline comparisons for our subsequent research question. 

To develop a fairer algorithm (\textbf{\textit{Research Question 2}}), we introduced a new predictive model based on a variational autoencoder (VAE), a probabilistic deep generative model~\citep{kingma2013auto}. 
Leveraging the VAE’s capability to extract latent features from input data, our approach was designed to disentangle sex-specific and sex-agnostic latent representations in the motion data. 
This disentanglement was intended to enable sex-fair predictions for box weights, even when the training data was sex-imbalanced. 
To contextualize our new algorithm, we first provide a brief overview of VAE before detailing the development of our sex-fair predictive model. 
We focus on key design insights while deferring the mathematical details to Appendix \ref{appendix}.

\subsubsection{Brief Overview of Variational Autoencoder (VAE)}\label{sec:VAE}

\begin{figure*}[t]
    \centering
    \includegraphics[width=0.9\textwidth]{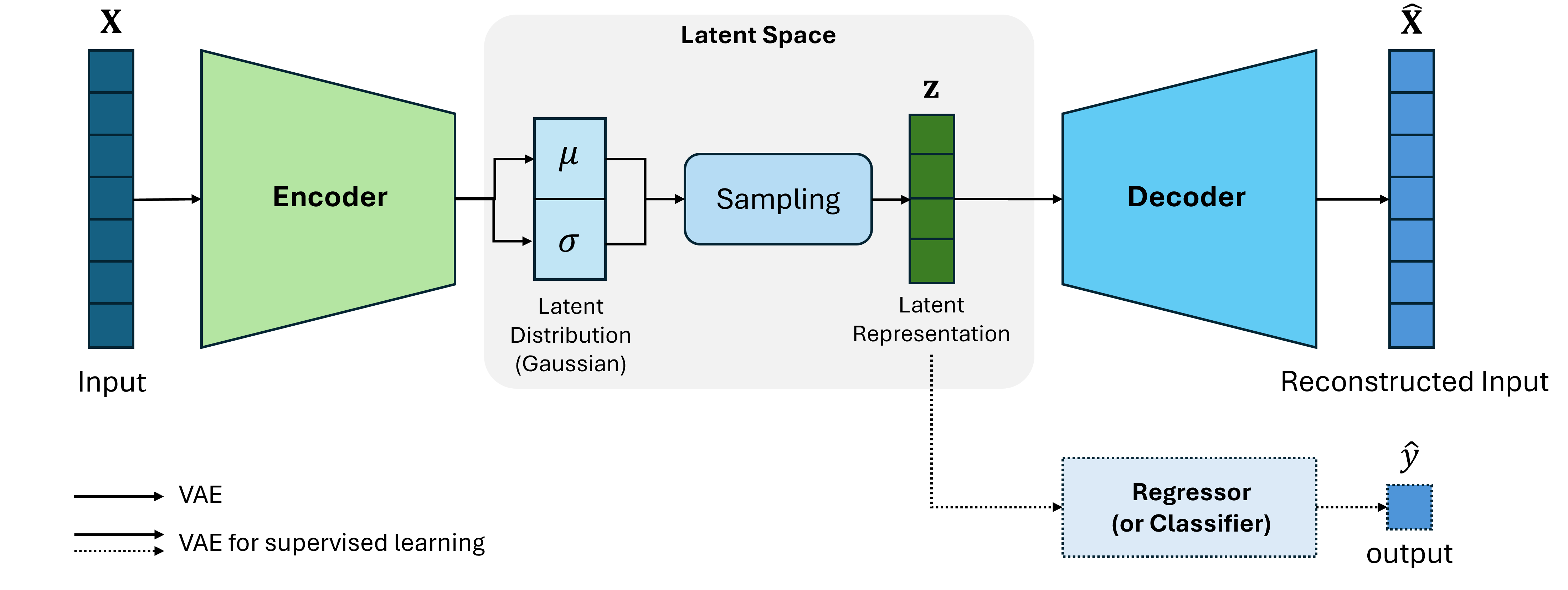}
    \caption{Structure of a Variational Autoencoder (VAE) and its extension for supervised learning. The latent distribution is often modeled as a Gaussian distribution with mean $\mu$ and standard deviation $\sigma$.}
    \label{fig:vae_vanilla}
\end{figure*}

The VAE was originally introduced as a probabilistic framework for learning efficient latent representations of data while enabling the generation of synthetic samples. Figure~\ref{fig:vae_vanilla} illustrates the structure of a VAE. It builds upon the structure of an autoencoder~\citep{li2023comprehensive}, which consists of two key components: an encoder and a decoder, both typically constructed using deep neural networks. The encoder maps input data into a lower-dimensional latent space, while the decoder reconstructs the original data from this latent representation. A key distinction of VAEs and traditional autoencoders is their probabilistic nature. Instead of mapping inputs to \textit{deterministic} latent representations, VAEs model the latent space using \textit{probability distributions}.
This allows the model to learn a latent variable distribution that captures the underlying structure of the data. As a result, VAEs can generate synthetic data by sampling from the learned latent distribution and decoding it into realistic outputs. Due to these capabilities, VAEs have been widely adopted across various domains for synthetic data generation, including computer vision~\citep{harvey2021conditional}, natural language processing~\citep{semeniuta2017hybrid}, biomedical applications~\citep{wei2020recent}, robotics~\citep{park2018multimodal}, and gait pattern analysis~\citep{larsen2024longitudinal}. Recently, VAE has also demonstrated significant success in \textit{supervised learning} tasks \citep[e.g.,][]{chamain2022end, yoo2017variational, zhao2019variational, berkhahn2019augmenting}. Our case, predicting box weights using sensor-based motion data, indeed falls within the domain of supervised learning. In such scenarios, VAEs are often extended to incorporate a classifier or regressor to map extracted latent representations to outputs (e.g., box weights). The effectiveness of VAEs in supervised learning is largely attributed to their ability to derive well-regularized latent representations from the inputs (e.g., sensor-based motion patterns)~\citep{jeon2021anomalous}. 
Motivated by the success of VAEs in supervised learning, we build upon this framework and redesign it to ensure fair predictions even in the presence of imbalanced training populations.




\subsubsection{Proposed Method: Debiasing VAE (\texttt{DVAE})} \label{sec:proposed_model}

\begin{figure*}[t]
    \centering
    \includegraphics[width=1\textwidth]{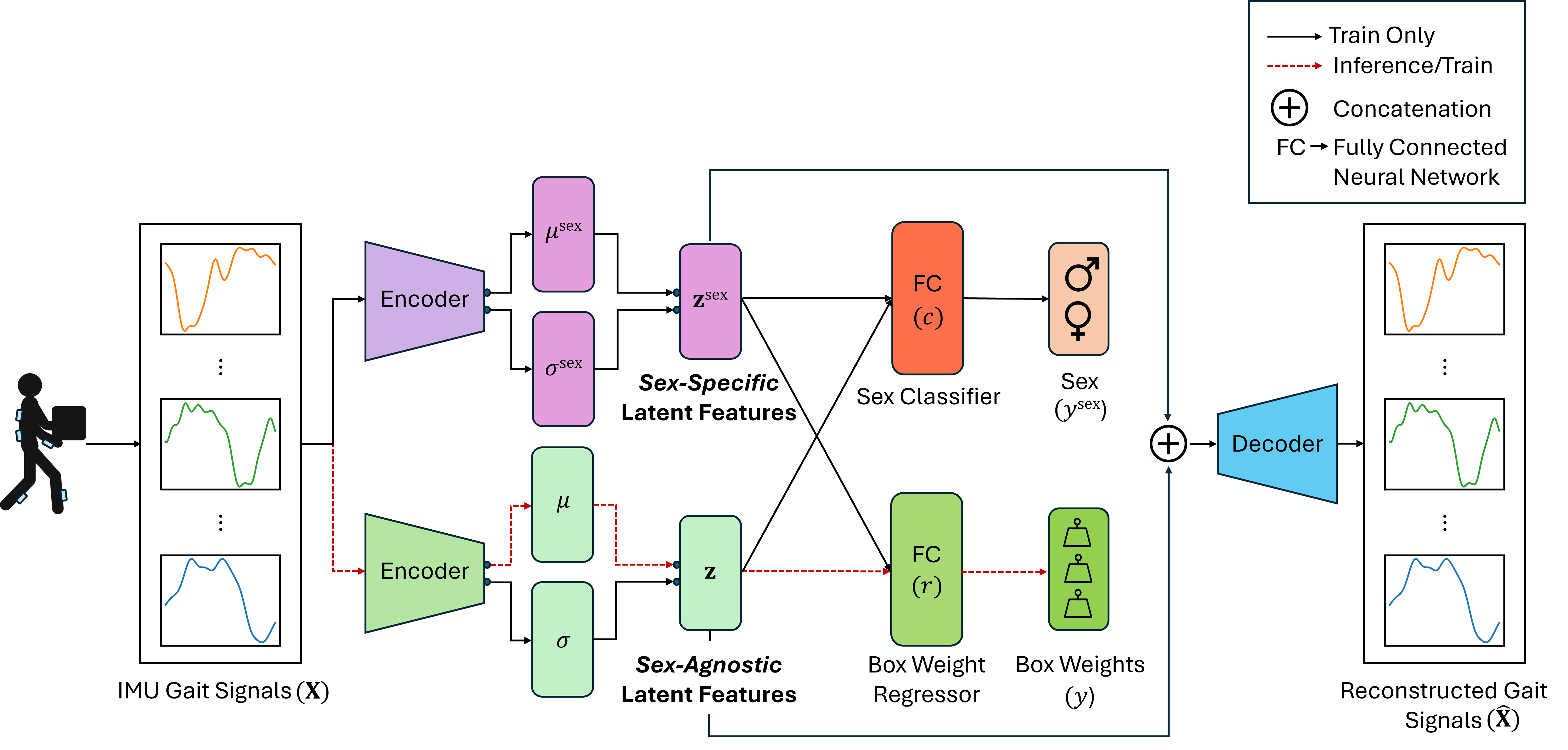}
    \caption{An overview of the \texttt{DVAE} model that separates the \textit{sex-specific} and \textit{sex-agnostic} features. During inference, $\sigma$ (standard deviation) is not used because the model directly utilizes $\mu$ (mean) for deterministic and point predictions, avoiding stochastic sampling.}
    \label{fig:overview}
\end{figure*}

We now introduce the design of our proposed Debiasing VAE (\texttt{DVAE}), specifically developed to enhance robustness against imbalanced populations. Figure~\ref{fig:overview} illustrates the overall architecture of our model. Inspired by latent independence excitation~\citep{qian2021latent}, which aims to disentangle latent features for domain generalization, our approach leverages the key intuition that learning \textit{sex-agnostic} latent features from motion data enables accurate box weight predictions regardless of whether the data is collected from male or female participants. 
Below, we outline the model architecture, training process, and inference procedure for test data, highlighting key insights at each stage. 
A more rigorous discussion of the mathematical formulations and derivations underlying our model is provided in Appendix~\ref{appendix}.

\paragraph{Model Architecture.} Our model adopts the encoder-decoder architecture of a VAE to construct a probabilistic latent space for embedding motion data. 
To facilitate fair predictions, we introduce a dual latent space design, separating latent representations into \textit{sex-agnostic} and \textit{\textit{sex-specific}} components. Specifically, our model employs two parallel encoders during the encoding process: one encoder extracts \textit{sex-specific} features, and the other captures \textit{sex-agnostic} features. 
The \textit{sex-agnostic} latent space represents motion patterns that are shared across sexes. The learned latent representations are then utilized for two downstream tasks: the \textit{sex-specific} latent space is linked to a neural network classifier that predicts sex, while the \textit{sex-agnostic} latent space is linked to a regressor that predicts box weight. The detailed architecture of the encoder and decoder is provided in Appendix~\ref{en}.

\paragraph{Model Training.} The training procedure of \texttt{DVAE} is performed by minimizing an average loss evaluated on training data. 
Given motion data for an arbitrary gait $\bX$, the corresponding participant's sex $y^{\sf sex}$, the box weight $y$, the loss function for a given gait $\ell(\bX; y^{\sf sex}, y)$, consists of three components and is formulated as: 
\begin{equation}\label{eq:loss}
    \ell(\bX; y^{\sf sex}, y) = \ell_{\sf VAE}(\bX) + \beta_1 \ell_{\sf DC}(\bX; y^{\sf sex}, y) +  \beta_2\ell_{\sf IE}(\bX; y, y^{\sf sex}),
\end{equation}
with the VAE loss $\ell_{\sf VAE}$, the discriminative loss $\ell_{\sf DC}$, the independence excitation loss $\ell_{\sf IE}$, and hyperparameters $\beta_1$ and $\beta_2$ that control the weights across losses. More specifically:

\begin{itemize}
    \item The VAE loss $\ell_{\sf{VAE}}$ evaluates the model’s reconstruction capability while regularizing the distribution of latent representations. Minimizing \(\ell_{\sf{VAE}}(\mathbf{X})\) encourages the formation of well-regularized latent spaces, where the embedding of the motion \(\mathbf{X}\) resides.
    
    \item The discriminative loss \(\ell_{\sf{DC}}\) assesses the predictive performance of the model in estimating both box weight \(y\) and sex \(y^{\sf{sex}}\) from the input motion data \(\mathbf{X}\). Minimizing \(\ell_{\sf{DC}} (\bX; y^{\sf sex}, y)\) facilitates the extraction of distinct latent representations: \textit{sex-agnostic} features that are crucial for box weight estimation and \textit{sex-specific} features for sex classification, while simultaneously optimizing the associated classifier $c$ and regressor $r$.
    
    \item The independence excitation loss \(\ell_{\sf{IE}}\) further enhances the disentanglement of \textit{sex-agnostic} and \textit{sex-specific} latent representations, aiming to ensure that \textit{sex-agnostic} features do not leak into the \textit{sex-specific} latent space and vice versa. This is achieved by \textit{weakening} the predictive ability of the classifier to infer sex $y^{\sf sex}$ based on \textit{sex-agnostic} latent feature $\bz$, as well as the regressor to estimate box weight $y$ based on \textit{sex-specific} latent feature $\bz^{\sf sex}$.
\end{itemize}

By jointly minimizing these three loss terms, the model learns well-regularized latent representations of human motion while effectively disentangling \textit{sex-agnostic} and \textit{sex-specific} features.

\paragraph{Inference.} During inference, we use only the \textit{sex-agnostic} encoder and the box weight prediction network (see red dotted arrows in Figure~\ref{fig:overview}). It begins with test inputs consisting of motion data of the same dimensionality as the training data, where each observation corresponds to a gait cycle. These inputs are processed through the \textit{sex-agnostic} encoder to extract sex-debiased features, which are then passed to the box weight prediction network to generate predictions for individual gait cycles. Since a single trial typically comprises multiple gait cycles, we compute the final prediction by averaging the predictions across all cycles within the trial. For a trial $j$ of participant $i$ with $n$ gaits, the predicted box weight can be written as $\hat{y}_{i,j} = \frac{1}{n} \sum_{k=1}^{n}\hat{y}_{i,j,k}$, where $\hat{y}_{i,j,k}$ is a predicted weight for the $k$-th gait. This averaged prediction $\hat{y}_{i,j}$ is then compared with the ground truth box weight.

\subsection{Model Performance Evaluation}\label{sec:model_eval} 
We trained ML models using training sets with varying male-to-female ratios to examine the impact of dataset composition on algorithmic biases. 
Five different ratios were used: 0.9:0.1, 0.7:0.3, 0.5:0.5, 0.3:0.7, and 0.1:0.9, representing a spectrum from male-dominant (0.9:0.1) to balanced (0.5:0.5) to female-dominant (0.1:0.9) training datasets. 
To assess the effects of these imbalances, we tested each model using male-only and female-only test sets, employing the Leave-One-Subject-Out Cross-Validation (LOSOCV) strategy. This approach enabled us to assess the impact of imbalanced training data on model performance and potential biases.

Model performance was assessed using mean absolute error (MAE) and three fairness metrics. To evaluate the impact of varying training datasets (male-to-female ratios) and test datasets (sex: male and female) on these performance metrics, separate two-way analyses of variance (ANOVAs) were used for each model and evaluation metric.
Significant effects were identified using $p<.05$, and post hoc paired differences were assessed using test slices.  
 Statistical analyses were performed using JMP Pro
v18.1.2 (SAS Institute, NC, USA).



To evaluate the ability to promote fairness, we assess models using three key fairness metrics: Statistical Parity (SP), Positive Residual Differences (PRD), and Negative Residual Differences (NRD). Each of these metrics is described below. Please note that we denote the actual and predicted box weights for participant $i$ as $y_i$ and $\hat y_i$, respectively, where we suppress the subscript $j$ for carrying condition for notational simplicity. 

\begin{itemize}
\item{\textbf{Statistical Parity (SP)}}: SP is a fairness metric used to assess whether a predictive model treats different demographic groups equitably~\citep{suarez2025general}. It measures the difference in the mean predicted outcomes between two groups—in this case, male and female test sets. SP is defined as:
\begin{equation}
\mathrm{SP} := \frac{1}{n_{f}} \sum_{i \in S_f}\hat{y}_i-\frac{1}{n_{m}} \sum_{i \in S_m}\hat{y}_i
\end{equation}
where \( S_f \) and \( S_m \) denote the female and male test sets, and \( n_{f} \) and \( n_{m} \) represent the number of female and male test samples, respectively. 

SP can be interpreted as a measure of whether a predictive distribution remains independent of the sensitive attribute, that is, biological sex in our case. A value of SP closer to 0 indicates that the model's predictions are more balanced between the groups, suggesting lower bias. Conversely, deviations from 0 imply that the model's performance differs across groups, indicating potential bias in the predictions.

\item{\textbf{Positive Residual Differences (PRD)}}: PRD measures the difference in underestimation errors between two demographic groups~\citep{johnson2022impartial}. PRD is defined as:
\begin{equation}
\mathrm{PRD}:=\left|\frac{1}{n_{f}} \sum_{i \in S_f} \max \left\{0, y_i-\hat{y}_i\right\}-\frac{1}{n_{m}} \sum_{i \in S_m} \max \left\{0, y_i-\hat{y}_i\right\}\right|
\end{equation}

PRD quantifies whether one group tends to have systematically higher positive residuals (i.e., actual values exceeding predicted values) compared to the other. Higher deviations from 0 indicate potential biases in model predictions.

\item{\textbf{Negative Residual Differences (NRD)}}: NRD measures the difference in overestimation errors between two demographic groups~\citep{johnson2022impartial}. NRD is defined as:
\begin{equation}
\mathrm{NRD}:=\left|\frac{1}{n_{f}} \sum_{i \in S_f} \min \left\{0, y_i-\hat{y}_i\right\}-\frac{1}{n_{m}} \sum_{i \in S_m} \min \left\{0, y_i-\hat{y}_i\right\}\right|
\end{equation}

NRD quantifies whether one group tends to have systematically higher negative residuals (i.e., predicted values exceeding actual values) compared to the other. A higher NRD value indicates a greater disparity in overestimation errors, suggesting potential bias in the model's predictions.

\end{itemize}

\section{Results}
\label{sec:results}

\begin{figure*}[!htbp]
    \centering
    \includegraphics[width=0.72\textwidth]{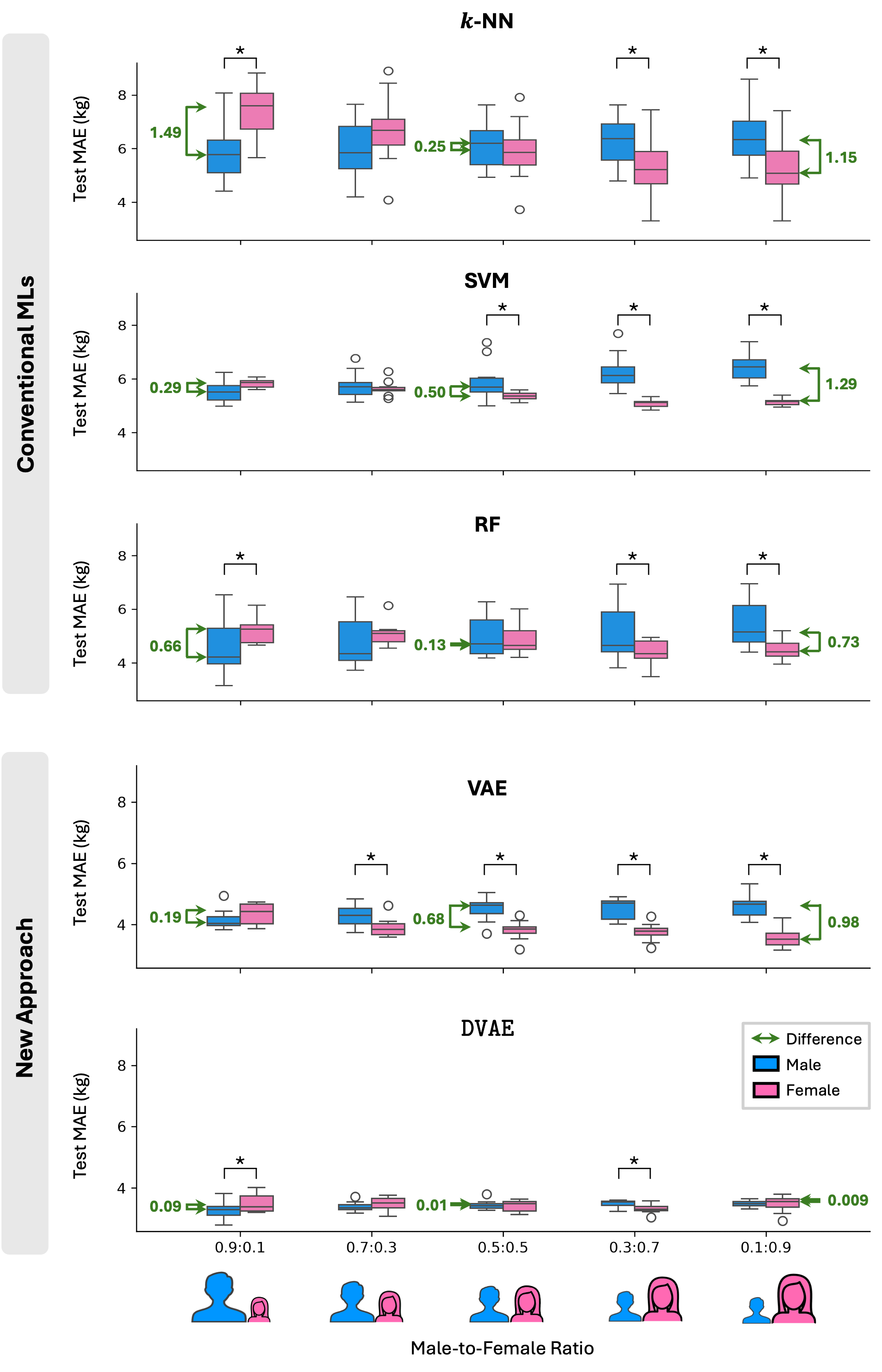}
    \caption{Performances of conventional ML models and new appraoches across varying male-to-female training ratios: Evaluation on male and female test sets, highlighting larger performance differences between groups in conventional MLs and VAE than \texttt{DVAE}. 
    The symbol “*” indicates a significant pairwise difference ($p < 0.05$).}
    \label{fig:box}
\end{figure*}

Figure~\ref{fig:box} presents boxplots illustrating the MAEs of predictions from conventional ML models ($k$-NN, SVM, and RF) and VAE-based approaches (VAE and \texttt{DVAE}). Specifically, it shows the MAEs for male and female test sets across varying male-to-female training ratios. Lower values indicate better performance (i.e., smaller errors). Table \ref{table:MLs_anova} summarizes the ANOVA results, examining the impact of two factors, i.e., the male-to-female ratio in the training dataset (the `Male-to-female ratio' factor) and sex groups (the `Sex' factor), on test MAEs, along with pairwise statistical comparisons of MAEs between male and female test groups for each male-to-female ratio in the training dataset. Based on these results, we below answer the two research questions raised in Section \ref{sec:intro}.

\begin{table}[h]
        \caption{Summary of ANOVA results [$F$ value, $p$ value] using MAE as dependent variable and Male-to-female ratio, Sex as independent variable for different models. Significant main and interaction effects are in bold font (\textit{p} $<$ .05).
        }
        ~\label{table:MLs_anova}
        \setlength\tabcolsep{8pt}
        \centering
        \begin{tabular}{lllll} 

        \toprule
         & Models  & Male-to-female ratio & Sex & Male-to-female ratio $\times$ Sex \\   

        \midrule
        \multirow{12}{*}{{Conventional ML models}} &
        $k$-NN & 2.12, .084 & 0.04, .849 & \textbf{5.70}, \textbf{$<$.001} \\
         &  &  &  & ~0.9:0.1, M $<$ F (10.02, .002)\\
         &  &  &  & ~0.3:0.7, M $>$ F (4.34, .040) \\ 
         &  &  &  & ~0.1:0.9, M $>$ F (5.99, .016) \\

        \cmidrule{2-5}
        & SVM & 0.50, .736 & \textbf{45.56, $<$.001} & \textbf{14.05}, \textbf{$<$.001} \\
         &  &  &  ~M $>$ F & ~0.5:0.5, M $>$ F (7.62, .007)\\
         &  &  &  & ~0.3:0.7, M $>$ F (40.79, $<$.001) \\ 
         &  &  &  & ~0.1:0.9, M $>$ F (50.50, $<$.001) \\

        \cmidrule{2-5}
        & RF & 0.32, .862 & 0.98, .325 & \textbf{4.35}, \textbf{.002} \\
         &  &  &  & ~0.9:0.1, M $<$ F (4.17, .044) \\
         &  &  &  & ~0.3:0.7, M $>$ F (4.40, .038) \\ 
         &  &  &  & ~0.1:0.9, M $>$ F (8.33, .005) \\

        \midrule

        \multirow{8}{*}{{New Approach}} &
        VAE & 0.88, .478 & \textbf{66.37, $<$.001} & \textbf{10.17}, \textbf{$<$.001} \\
         &  &  & ~M $>$ F & ~0.7:0.3, M $>$ F (6.74, .011)\\
         &  &  &  & ~0.5:0.5, M $>$ F (22.55, <.001) \\ 
         &  &  &  & ~0.3:0.7, M $>$ F (28.99, <.001) \\
         &  &  &  & ~0.1:0.9, M $>$ F (46.92, <.001) \\
         
        \cmidrule{2-5}
        & \texttt{DVAE} & 0.85, .496 & 0.50, .483 & \textbf{3.17}, \textbf{.017} \\
         &  &  &  & ~0.9:0.1, M $<$ F (7.64, .007) \\
         &  &  &  & ~0.3:0.7, M $>$ F (4.11, .045) \\ 
         
        \bottomrule
        \end{tabular}
        \end{table}

\subsection{\textit{Research Question 1}: Do Conventional ML Algorithms Exhibit Bias When Predicting Hand Load?}
\label{sec:bias}


From the top three plots in Figure \ref{fig:box}, we derive key insights in respect to Research Question 1. Notably, it is clear to see that all conventional ML models exhibit significant disparities in MAE across different sexes. For $k$-NN and RF, statistically substantial performance gaps (indicated by ``$*$'') emerge when trained on highly imbalanced populations, either male-dominant (0.9:0.1) or female-dominant (0.1:0.9). The most pronounced MAE difference is observed for $k$-NN at the 0.9:0.1 ratio, reaching MAE of 1.49. Meanwhile, performance disparities are minimized at the balanced 0.5:0.5 ratio for both $k$-NN and RF, as a balanced dataset enables the model to learn robust feature representations from both female and male samples, facilitating better generalization. Compared to RF and $k$-NN, SVM exhibits a stronger bias towards female groups. Even when trained on a balanced (0.5:0.5) dataset, SVM shows a significant performance disparity favoring the female test set, with a mean difference of 0.50. Across all conventional ML models, predictions for the female test set improve as the proportion of females in the training set increases, and a similar trend is observed for male predictions when male proportion increases. This highlights the inherent nature of data-driven models, which learn directly from the provided data, and thus emphasizes the need for mitigation strategies to achieve fair predictions rather than blindly applying ML models to imbalanced training populations.

A more rigorous statistical analysis of MAEs is provided in Table~\ref{table:MLs_anova}. The two-way ANOVA results indicate that, for all conventional ML models, there are statistically significant interaction effects between the male-to-female ratio and sex in the training population. In other words, the impact of training ratio on test MAEs substantially differs between male and female test groups. This aligns with the patterns observed in Figure \ref{fig:box}, where an increasing proportion of females in the training set leads to lower MAEs for female test samples but higher MAEs for male test samples, for all conventional ML models. Another notable observation is that the main effect of the Sex factor on SVM's MAEs is statistically significant, where female MAEs are substantially lower than male MAEs ({\sf M $>$ F}). This outcome suggests that SVM predictions overall exhibit a bias toward females, consistent with our previous observations in Figure \ref{fig:box}.

\subsection{\textit{Research Question 2}: If Bias is Present, Can We Mitigate It by Developing a Fair ML Algorithm?}\label{sec:result_RQ2}
\label{sec:model_performance}
Given the significant algorithmic biases observed in conventional ML models, we address Research Question 2 by comparing the performance of our proposed debiasing model (\texttt{DVAE}) against the baseline VAE model and the three conventional ML models. The comparison is based on both MAEs and fairness metrics introduced in Section \ref{sec:model_eval}.


\paragraph{Model Comparison Using MAE.}
The bottom two plots in Figure~\ref{fig:box} present the MAE results for the baseline VAE and our proposed debiasing model \texttt{DVAE}. Here, we clearly observe that \texttt{DVAE} consistently exhibits lower MAE deviation between female and male test sets across all training ratios compared to both conventional models and VAE, demonstrating its strong bias mitigation capability even when trained on imbalanced populations. Notably, this improvement is achieved alongside superior predictive performance, as reflected in the lowest average MAE and standard deviation among all compared models. In particular, the reduced standard deviation highlights \texttt{DVAE}’s robustness across different cross-validation sets, surpassing other benchmarks. In contrast, the standard VAE model, which lacks explicit bias mitigation mechanisms, exhibits significant prediction bias, particularly favoring female test samples. Overall, VAE-based approaches outperform conventional ML models in predictive accuracy, but only \texttt{DVAE} effectively balances both fairness and performance.

As shown in the bottom two major rows of Table \ref{table:MLs_anova}, the $F$ value for the interaction effect in \texttt{DVAE} (3.17) is substantially lower than those of VAE and conventional ML models. This suggests that the influence of sex on the impact of varying training ratio on predictive performance is significantly reduced, demonstrating \texttt{DVAE}’s enhanced bias mitigation capability. Indeed, the reduced interaction effect aligns with Figure \ref{fig:box}, where the deviation in MAE trends along different training ratios for male and female groups is notably mitigated in \texttt{DVAE} compared to other benchmark models. Lastly, the significant main effect of the sex attribute in VAE confirms its prediction bias, which systematically favors female samples.

\begin{figure*}
    \centering
    \includegraphics[width=0.9\textwidth]{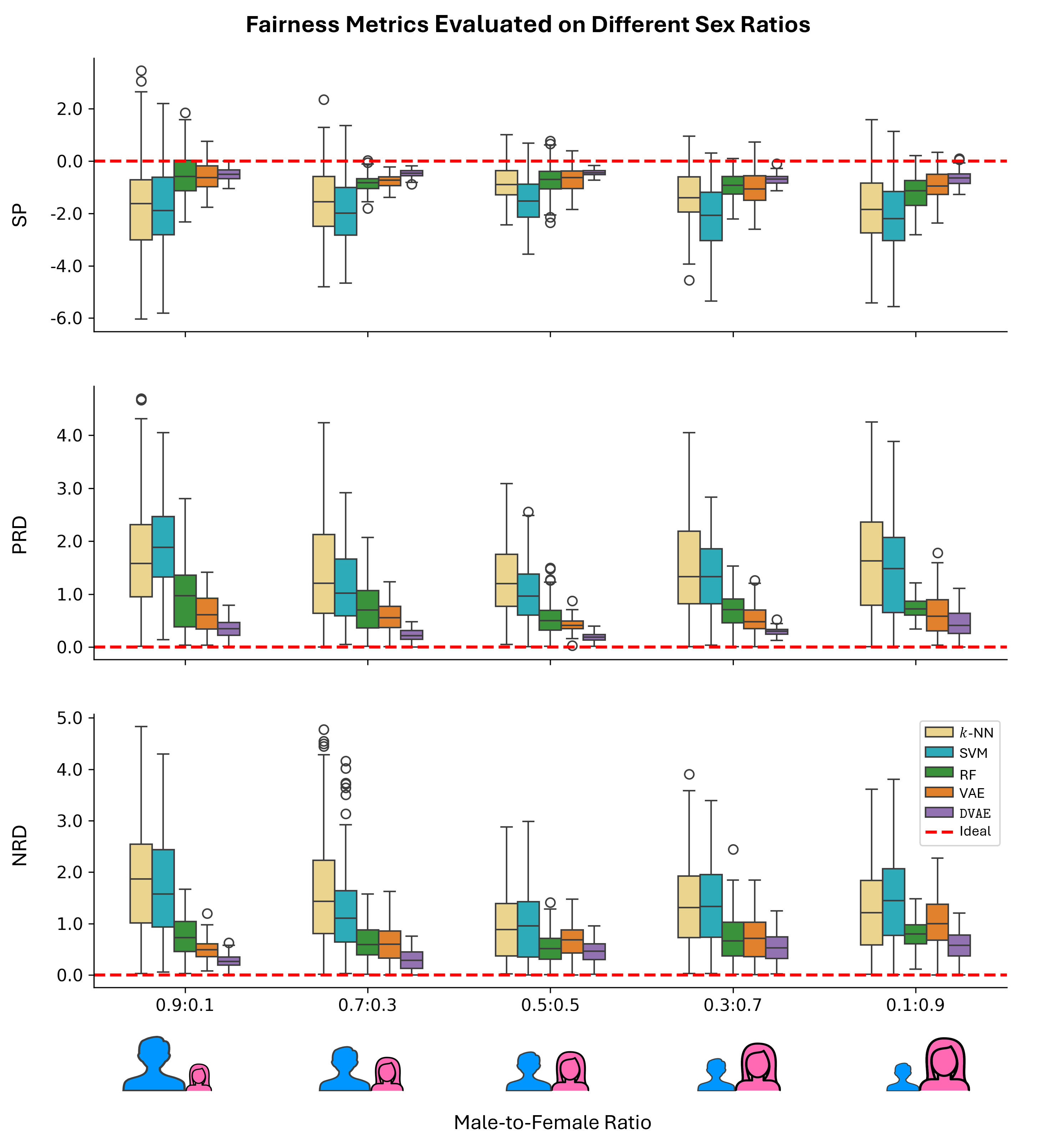}
    \caption{Fairness metrics evaluated on both female and male test sets across different male-to-female training ratios. Values closer to the dotted red line indicate greater fairness based on the selected fairness metric. SP = Statistical Parity, PRD = Positive Residual Differences, NRD = Negative Residual Differences.}     
    \label{fig:fairness}
\end{figure*}

\paragraph{Model Comparison Using Fairness Metrics.}
To further assess the fairness of each model beyond MAE, we computed three fairness metrics: SP, PRD, and NRD, as discussed in Section \ref{sec:model_eval}. Figure~\ref{fig:fairness} presents these metrics for all five models for comparison. The red dotted line represents the ideal value for each fairness metric, providing a reference for evaluating model fairness.

\begin{table}
\centering
\setlength\tabcolsep{3pt}
\caption{Summary of ANOVA results [$F$ value, $p$ value] using SP, PRD, and NRD separately as dependent variable and Male-to-female ratio, Model as independent variable. Significant main and interaction effects are in bold font (\textit{p} $<$ .05).}
\label{table:fairness_anova}
\begin{tabular}{llll} 
\toprule
Fairness metric      & Male-to-female ratio    & Model                    & Male-to-female ratio $\times$ Model\\ 
\midrule
\multirow{6}{*}{SP}  & \textbf{25.87, $<$.001} & \textbf{233.88, $<$.001} & \textbf{5.26, $<$.001}\\
                     & ~0.5:0.5 $>$ 0.9:0.1 $>$ & ~\texttt{DVAE} $>$ VAE $>$ & ~0.9:0.1,~\texttt{DVAE} $>$ RF $>$ VAE $>$ $k$-NN $>$ SVM (72.24, $<$.001) \\
                     & ~0.7:0.3 $>$ 0.3:0.7 $>$ & ~RF $>$ $k$-NN $>$   & ~0.7:0.3,~\texttt{DVAE} $>$ VAE $>$ RF $>$ $k$-NN $>$ SVM (53.18, $<$.001) \\
                     & ~0.1:0.9              & ~SVM     & ~0.5:0.5,~\texttt{DVAE} $>$ VAE $>$ RF $>$ $k$-NN $>$ SVM (26.47, $<$.001) \\
                     &                       &          & ~0.3:0.7,~\texttt{DVAE} $>$ RF $>$ VAE $>$ $k$-NN $>$ SVM (46.99, $<$.001) \\
                     &                       &          & ~0.1:0.9,~\texttt{DVAE} $>$ VAE $>$ RF $>$ $k$-NN $>$ SVM (56.04, $<$.001) \\ 
\hline
\multirow{6}{*}{PRD} & \textbf{44.37, $<$.001}  & \textbf{462.37, $<$.001}    & \textbf{5.01, $<$.001}\\
                     & ~0.9:0.1 $>$ 0.1:0.9 $>$ & ~$k$-NN $>$ SVM $>$ & ~0.9:0.1,~SVM $>$ $k$-NN $>$ RF $>$ VAE $>$ \texttt{DVAE} (147.29, $<$.001)\\
                     & ~0.3:0.7 $>$ 0.7:0.3 $>$  & ~RF $>$ VAE $>$  & ~0.7:0.3,~$k$-NN $>$ SVM $>$ RF $>$ VAE $>$ \texttt{DVAE} (73.77, $<$.001) \\
                     & ~0.5:0.5               & ~\texttt{DVAE}      & ~0.5:0.5,~$k$-NN $>$ SVM $>$ RF $>$ VAE $>$ \texttt{DVAE} (68.25, $<$.001) \\
                     &                        &                     & ~0.3:0.7,~$k$-NN $>$ SVM $>$ RF $>$ VAE $>$ \texttt{DVAE} (91.84, $<$.001) \\
                     &                        &                     & ~0.1:0.9,~$k$-NN $>$ SVM $>$ RF $>$ VAE $>$ \texttt{DVAE} (101.26, $<$.001) \\ 
\hline
\multirow{6}{*}{NRD} & \textbf{24.28, $<$.001}  & \textbf{299.32, $<$.001}    & \textbf{14.55, $<$.001}\\
                     & ~0.1:0.9 $>$ 0.9:0.1 $>$ & ~$k$-NN $>$ SVM $>$  & ~0.9:0.1,~$k$-NN $>$ SVM $>$ RF $>$ VAE $>$ \texttt{DVAE} (161.98, $<$.001)\\
                     & ~0.3:0.7 $>$ 0.7:0.3 $>$   & ~VAE $>$ RF $>$    & ~0.7:0.3,~$k$-NN $>$ SVM $>$ RF $>$ VAE $>$ \texttt{DVAE} (91.74, $<$.001)\\
                     & ~0.5:0.5               & ~\texttt{DVAE}  & ~0.5:0.5,~SVM $>$ $k$-NN $>$ VAE $>$ RF $>$ \texttt{DVAE} (17.58, $<$.001)\\
                     &                        &                 & ~0.3:0.7,~SVM $>$ $k$-NN $>$ RF $>$ VAE $>$ \texttt{DVAE} (47.13, $<$.001)\\
                     &                        &                 & ~0.1:0.9,~SVM $>$ $k$-NN $>$ VAE $>$ RF $>$ \texttt{DVAE} (39.10, $<$.001)\\
\bottomrule
\end{tabular}
\end{table}



From Figure \ref{fig:fairness} we derive several key insights. 
First, compared to other models, \texttt{DVAE} consistently achieves SP, PRD, and NRD values that are substantially closer to the ideal across different male-female ratios in the training data. 
The near-zero SP values of \texttt{DVAE}, relative to other models, indicate that \texttt{DVAE} produces less biased predictions, avoiding systematic favoritism toward one sex. 
Similarly, the near-zero PRD and NRD values suggest that when \texttt{DVAE} does overestimate or underestimate box weights, it does so equitably across male and female groups. 
Second, the fairness metrics of \texttt{DVAE} exhibit substantially lower variance than those of other models across all male-female ratios. This reduced variance demonstrates that \texttt{DVAE} maintains a more consistent level of predictive fairness regardless of the sex composition of the training data. Third, fairness metrics generally improve as the male-female ratio approaches balance (0.5:0.5) across all models. This trend is expected, as fair predictions are easier to achieve when training data is more representative of both groups. Notably, the improvement is observed not only in the median values of the metrics but also in their variance, indicating that balanced training datasets lead to more consistently fair predictions, whereas imbalanced datasets result in both greater variability and deterioration in fairness. Finally, VAE does not always outperform RF in terms of the fairness metrics considered, implying that the improved prediction accuracy of VAE over RF (as discussed in Section \ref{sec:result_RQ2}) does not necessarily indicate better fairness. This indeed highlights the importance of considering fairness in addition to predictive accuracy when developing ML algorithms for ergonomic risk exposure.


Table~\ref{table:fairness_anova} presents a summary of two-way ANOVA results showing the comparisons of fairness across conventional ML models and our proposed approaches. 
We found significant interaction effects in all fairness metrics, suggesting that the choice of prediction model significantly influences how the varying training ratios affect fairness metrics.
For SP, \texttt{DVAE} achieved the largest SP value among models, which is the closest value to zero (ideal) given that every model shows negative median SP values (see Figure \ref{fig:fairness}). Likewise, \texttt{DVAE} exhibits the smallest PRD and NRD, both of which are non-negative by definition, again indicating its proximity to the ideal value zero.
The significance of the main effect of the Male-to-female ratio factor suggests that, on average across all models, the metric value differences between different training ratios are statistically significant. These differences are such that a 0.5:0.5 training ratio results in the fairest predictions, achieving the highest SP and the lowest PRD and NRD. This finding demonstrates that ML models generally achieve better fairness when trained on a balanced population, and conversely, fairness decreases with imbalanced populations.

\section{Discussion}
\label{sec:discussion}
Our findings indicate that commonly used ML algorithms for hand load estimation exhibit systematic biases when trained on sex-imbalanced datasets. However, our proposed \texttt{DVAE} approach effectively mitigates these biases. In the following section, we discuss the implications of these findings, emphasizing bias mitigation strategies and practical applications while outlining our suggested future directions.

\subsection{VAE-based Models vs. Conventional ML Models}
Our study demonstrated that deep generative models, such as VAE, significantly outperform traditional machine learning models in estimating box weight. 
When comparing MAEs across different training ratios and sexes, we observed a clear improvement: $k$-NN performed the worst (MAE = 6.13), followed by RF (MAE = 4.89), then VAE (MAE = 4.17), with our proposed \texttt{DVAE} achieving the best accuracy (MAE = 3.42).
This improvement comes from VAE’s ability to learn rich representations from complex IMU data and its probabilistic nature, which makes it more robust to movement noise. Our findings align with previous research showing that deep generative models are particularly effective for human motion analysis~\citep{li2021lifting, qing2024predicting}, highlighting their strength in capturing complex relationships between movement data and external loads.
Beyond accuracy, our proposed \texttt{DVAE} also outperformed both conventional ML models and the baseline VAE in fairness metrics. These results indicate that \texttt{DVAE} is not only the most accurate model but also the most consistent and equitable among those we evaluated.

\subsection{Reducing Bias: How \texttt{DVAE} Separates Sex-Specific and Task-Relevant Features}

\begin{figure*}
    \centering
    \includegraphics[width=0.7\textwidth]{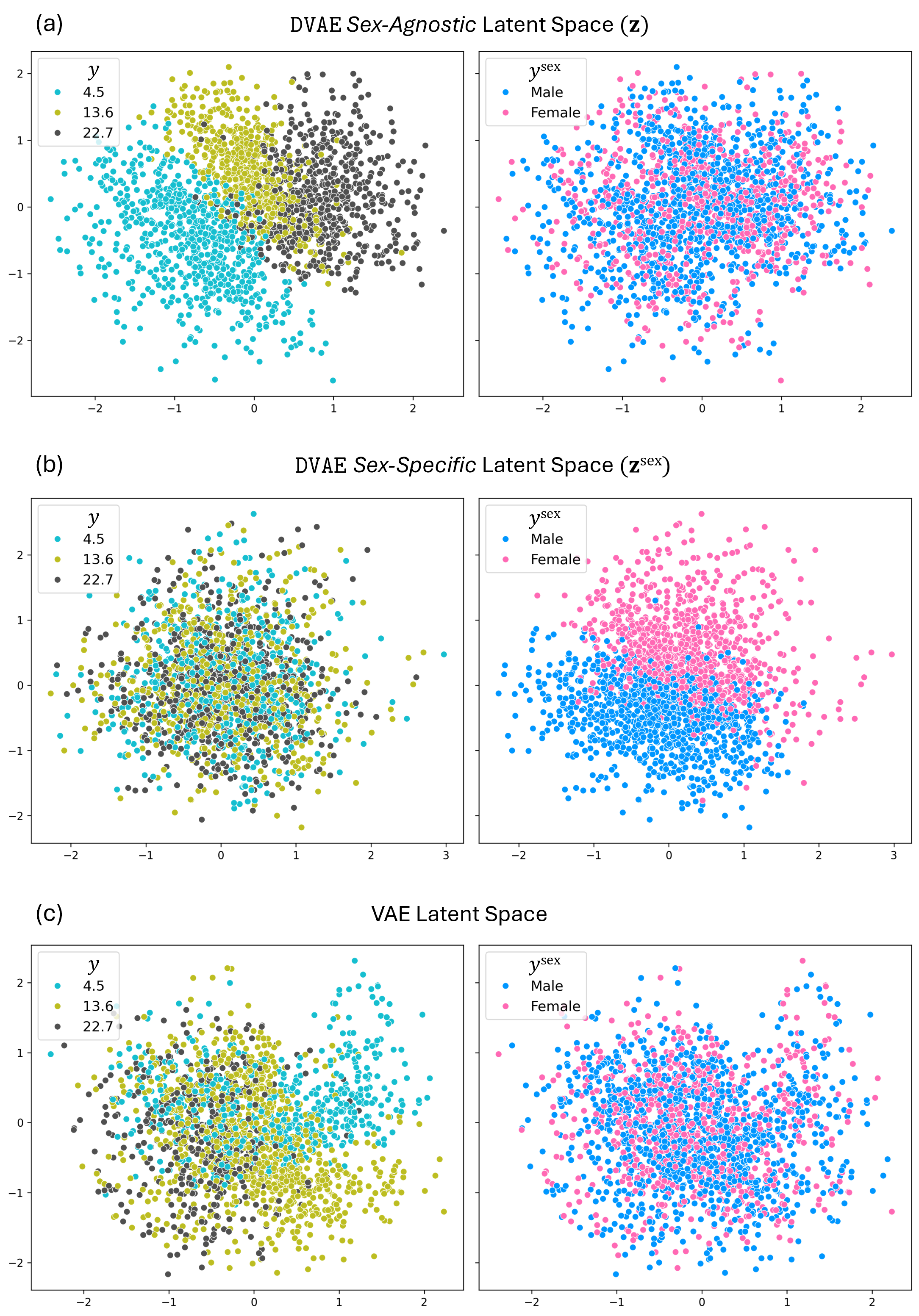}
    \caption{Latent space visualization: (a) \texttt{DVAE} \textit{sex-agnostic} latent space, showing clear separation between different box weights while ignoring sex differences; (b) \texttt{DVAE} \textit{sex-specific} latent space, showing a clear separation between male and female while ignoring box weights; and (c) VAE, showing no clear separation in either box weights or sexes.}
    \label{fig:latent}
\end{figure*}

Our \texttt{DVAE} approach successfully mitigated sex-based bias by separating sex-specific features from the key patterns needed to estimate hand load. To better understand how this worked, we visualized the latent features learned by both \texttt{DVAE} and a standard VAE using a technique called t-distributed Stochastic Neighbor Embedding (t-SNE)~\citep{van2008visualizing}, which helps display complex data in a simple 2D space (Figure~\ref{fig:latent}). 
Figure~\ref{fig:latent}(a) shows the \texttt{DVAE}'s "sex-agnostic" features, where data points are grouped by box weight but not by sex, meaning the model focuses on weight rather than sex differences.
In contrast, Figure~\ref{fig:latent}(b) shows the "sex-specific" features, where data is separated by sex but not by box weight, confirming that \texttt{DVAE} successfully isolates sex-related information.
Figure~\ref{fig:latent}(c) illustrates the results from a standard VAE, which struggles to separate sex and weight-related features. Here, sex and weight boundaries are mixed together, meaning the model may unintentionally use sex information when predicting box weights. This comparison highlights how \texttt{DVAE} effectively prevents unwanted bias by ensuring that only relevant, sex-agnostic movement patterns are used for box weight estimation. As a result, our approach enables fairer and more accurate predictions across different sex groups.


\subsection{Practical Implications and Recommendations}
Currently, over two dozen commercial systems use machine learning and artificial intelligence to assess worker exposure risks. These systems estimate risk scores or identify exposure to ergonomic risk factors based on worker posture data. Many claim to incorporate advanced algorithms that go beyond posture analysis—such as considering factors like box weights, as investigated in our study—by extracting contextual information about the work environment and tasks.
However, the proprietary nature of these algorithms raises concerns about fairness and potential biases across worker populations. While algorithm-driven ergonomic assessment tools hold significant promise for improving workplace safety, systematic biases in these models—particularly across different demographic groups—could lead to inequitable risk assessments and misinformed interventions.

In the field of human factors and ergonomics, an increasing number of studies are adopting machine learning methods to enhance worker risk assessment. 
However, ensuring fairness and transparency in these models remains a critical yet unaddressed challenge.
Our study serves as an important first step in examining algorithmic biases, particularly those related to biological sex, in ergonomic assessment models.
Based on our findings, we offer the following recommendations for researchers and program developers to consider when designing data-driven ergonomic assessment algorithms:

\paragraph{Selecting the Right Prediction Model.}
While our study focused on sex-based bias as a case study, our methodology can be extended to address biases related to other sensitive attributes. For example, the relationship between IMU-based gait patterns and external load can vary significantly depending on factors beyond biological sex, such as age, anthropometry, strength, and prior work experience. When the training dataset is unbalanced with respect to these attributes, algorithmic bias may emerge.

Although we cannot guarantee how well our approach (\texttt{DVAE}) mitigates algorithmic biases while maintaining prediction accuracy when used with other attributes, our findings highlight a promising direction for improving fairness using enhanced deep generative models specifically designed to address bias. 
Specifically, for future researchers interested in testing \texttt{DVAE}, our method can be easily adapted by replacing the categorical label $y^{\sf sex}$ in Figure~\ref{fig:overview} with another label corresponding to the sensitive attribute of interest, such as age, strength, or other demographic factors.

Additionally, other machine learning techniques have shown potential in mitigating biases. Methods such as adversarial debiasing and transfer learning have been tested in health-related wearable applications, including Parkinson’s disease monitoring, and could be explored further to enhance fairness in ergonomic risk assessment models. For instance,~\citet{odonga2025bias} demonstrated that transfer learning from multi-site and generic human activity datasets significantly improved both fairness and performance in detecting freezing of gait. Likewise,~\citet{zhu2023wearable} showed that integrating a Multi-Attribute Fairness Loss into convolutional neural network (CNN) architectures outperformed several baseline fairness-aware methods in wearable-based pain assessment, particularly by reducing disparities across race, gender, and cognitive ability.


\paragraph{Selecting the Right Fairness Metric.}
Our evaluation using multiple fairness metrics—SP, PRD, and NRD—provided a comprehensive perspective on model fairness. SP measures how strongly the prediction distribution is influenced by a sensitive attribute (in our case, biological sex). PRD and NRD complement SP by highlighting differences in overestimation and underestimation errors between sexes.
Since a model that appears fair under one metric may still exhibit bias under another \citep{kleinberg2016inherent}, it is crucial to consider multiple fairness metrics rather than relying on a single criterion. By analyzing how different ML models performed across SP, PRD, and NRD at various training ratios, future studies using ML algorithms can gain a more nuanced understanding of algorithmic bias and fairness for their applications.

\subsection{Limitations and Future Directions}
Our study has several limitations worth discussing. 
While our model demonstrates strong performance in controlled laboratory conditions with a relatively small sample size (22 participants), its effectiveness in real-world industrial settings---where worker populations are more diverse and environmental conditions vary---remains to be validated.
Additionally, our analysis focused solely on sex as a demographic factor, whereas other important attributes, such as age, anthropometry, and prior work experience, could also influence movement patterns and load-carrying capabilities.
Future research should explore these factors to better understand their impact on algorithmic fairness and model performance. 
Moreover, the generalizability of our approach should be tested in more complex real-world scenarios, including dynamic load conditions and varying terrains. Investigating methods such as few-shot learning \citep{finn2017model} or real-time adaptive frameworks \citep{chung2025real} could help develop a more responsive and fine-tuned model capable of adapting to diverse occupational settings and worker populations, while maintaining the fairness of the algorithm output. 

\section{Conclusions}
\label{sec4}
This paper investigates algorithmic biases in machine learning models used to predict hand-carried box weights based on IMU sensor gait patterns. We found that commonly used ML models can introduce bias when trained on sex-imbalanced datasets, leading to unfair predictions across different sex groups. To address this issue, we developed Debaising VAE (\texttt{DVAE}), a model designed to reduce bias by separating sex-agnostic and sex-specific features in gait patterns. By ensuring that weight predictions rely only on sex-agnostic features, \texttt{DVAE} makes fairer predictions for both biological sexes.
Compared to conventional ML models like $k$-NN (MAE = 6.13) and Random Forest (MAE = 4.89), deep generative models performed significantly better, with VAE achieving an MAE of 4.17 and our proposed \texttt{DVAE} achieving the best accuracy (MAE = 3.42). Additionally, \texttt{DVAE} outperformed other models in three fairness metrics (SP, PRD, NRD), demonstrating its ability to provide both more accurate and fairer predictions.
These results show that \texttt{DVAE} not only improves prediction accuracy but also enhances fairness, making it a promising approach for reducing bias in ergonomic assessments.


\section*{Acknowledgement}
This research is supported by the National Science Foundation (Grant No. 2427599) and the National Safety Council through the Research to Solutions (R2S) Grant awards.

\section*{Disclosure Statement}
No potential conflict of interest was reported by the authors.



\bibliographystyle{apalike}

\bibliography{references}


\newpage
\appendix
\section{Appendix}\label{appendix}

\renewcommand{\theequation}{A\arabic{equation}}  
\setcounter{equation}{0}  

\subsection{Technical Details for \texttt{DVAE}}

\texttt{DVAE} models the joint distributions \(\mathbb{P}^d(\bX, y)\) of the input $\bX$ (e.g., IMU signals) and 
the output $y$ (e.g., box weights), where $d$ represents the domain index (e.g., male and female). These distributions are distinct while sharing the same label space $y$. During testing, target domain data comes from unseen participants. The objective is to train a function $f: \bX \rightarrow y$ that generalizes effectively across domains.



\texttt{DVAE} learns to extract a latent representation from input $\bX$. Unlike standard VAEs, \texttt{DVAE} distinguishes itself by decomposing the latent space $(\bz, \bz^{\sf sex})$, where $\bz$ captures sex-agnostic features and $\bz^{\sf sex}$ captures sex-specific features. Following the VAE framework, we define the probabilistic encoders as:
\begin{align}
    &q(\bz|\bX; \bpsi) := \cN(\bmu_\bpsi(\bX), \bSigma_\bpsi(\bX)),\label{A1}\\
    &q(\bz^{\sf sex} | \bX; \bpsi^{\sf sex}) := \cN(\bmu_{\bpsi^{\sf sex}}(\bX), \bSigma_{\bpsi^{\sf sex}}(\bX))\label{A2}.
\end{align}
Both encoders are modeled as Gaussian distributions, where their means and diagonal covariances are parameterized by neural networks that take $\bX$ as input, with parameters $\bpsi$ and $\bpsi^{\sf sex}$, respectively. We also define the probabilistic decoder $p(\bX|\bz,\bz^{\sf sex}; \bphi)$ expressed as:
\begin{equation}
    p(\bX|\bz,\bz^{\sf sex}; \bphi) := \cN(\mu_\bphi(\bz,\bz^{\sf sex}), \sigma^2_{\bphi}(\bz,\bz^{\sf sex}))\label{A3},
\end{equation}
where $\mu_\bphi(\bz,\bz^{\sf sex})$ and $\sigma^2_{\bphi}(\bz,\bz^{\sf sex})$ are the mean and variance of the Gaussian distribution, again parameterized by a neural network with $\bphi$ that takes both $\bz$ and $\bz^{\sf sex}$ as input. Given \eqref{A1}-\eqref{A3}, the parameters $\{\bpsi, \bpsi^{\sf sex}, \bphi\}$ in the encoders and the decoder are jointly optimized by minimizing the VAE loss written as:
\begin{align}
\ell_{\sf VAE}(\bX) = & -\mathbb{E}_{q\left(\bz^{\sf sex} \mid \bX ; \bpsi^{\sf sex}\right) q\left(\bz \mid \bX ; \bpsi\right)} \left[\log p\left(\bX \mid \bz, \bz^{\sf sex} ; \bphi\right)\right] \nonumber\\
&+{\sf KL}\left(q\left(\bz^{\sf sex} \mid \bX ; \bpsi^{\sf sex}\right) \| p\left(\bz^{\sf sex}\right)\right)
+{\sf KL}\left(q\left(\bz \mid \bX ; \bpsi\right) \| p(\bz)\right)\label{A4}
\end{align}
with the priors $p(\bz)$ and $p\left(\bz^{\sf sex}\right)$ being standard Gaussians and ${\sf KL}(\cdot \Vert \cdot)$ indicating the Kullback-Leibler divergence between two probability distributions. We refer the reader to \cite{kingma2013auto} for the theoretical background of VAE to derive \eqref{A4}.

The discriminative loss $\ell_{\sf DC}$ in \eqref{eq:loss} is related to two separate networks incorporated into the model: a regressor $r$ and a classifier $c$, with parameters $\mathbf{w}$ and $\mathbf{w}^{\sf sex}$, respectively. The neural network $r(\cdot; \mathbf{w})$ is trained to predict box weight $y$ based on $\bz$, while $c(\cdot; \mathbf{w}^{\sf sex})$ is trained to predict the participant's sex $y^{\sf sex}$ using $\bz^{\sf sex}$. The loss function of this discriminative network, applied to a single sample, is defined as:
\begin{equation}
    \ell_{\sf DC}(\bX; y^{\sf sex}, y) = 
    \ell_r\left(y, r\left(\bz; \mathbf{w}\right)\right) 
    +\ell_c\left(y^{\sf sex}, c\left(\bz^{\sf sex}; \mathbf{w}^{\sf sex}\right)\right)\label{A5}
\end{equation}
with task-specific loss functions $\ell_r$ and $\ell_c$, such as cross-entropy or mean squared error (MSE). 

The independence excitation loss $\ell_{\sf IE}$ in \eqref{eq:loss} is designed to \textit{compromise} the performance of the sex classifier $c$ when provided with sex-agnostic latent feature $\bz$, and at the same time, compromise the performance of the box weight regressor $r$ when provided with $\bz^{\sf sex}$, expressed as: 
\begin{equation}\label{A6}
    \ell_{\sf IE}(\bX; y, y^{\sf sex}) = 
    -\ell_r\left(y, r\left(\bz^{\sf sex}; \mathbf{w}\right)\right) 
    -\ell_c\left(y^{\sf sex}, c\left(\bz ; \mathbf{w}^{\sf sex}\right)\right),
\end{equation}

Finally, the losses in \eqref{A4}, \eqref{A5}, and \eqref{A6} together form the overall loss of \texttt{DVAE}, as defined in \eqref{eq:loss}.

\section{Encoder-Decoder Architecture and Hyperparameters}
\label{en}

We designed an encoder-decoder architecture using a 1D Convolutional Neural Network (CNN). The details of the architecture are as follows:

\begin{itemize}
    \item The encoder begins with a 1D CNN with an input dimension of $128 \times 72$, followed by three convolutional layers (with 64, 128, and 256 filters, respectively), each followed by MaxPooling (kernel size = 2) for downsampling. The output is flattened and passed through a fully connected layer (dimension: $256 \times 128$), ReLU activation, followed by another linear layer ($128 \times 64$), and a final linear layer ($64 \times \text{latent\_dim} = 16$) to generate the mean and log-variance for latent space sampling. 
    
    \item The decoder mirrors the encoder structure, using a fully connected layer ($16 \times 64$), followed by layers expanding back to 128 and 256 dimensions, with upsampling and transpose convolutions to reconstruct the original signal. 
    
    \item For classification, we used a fully connected neural network with an input dimension of 16, followed by two linear layers ($16 \times 128$ and $128 \times 64$), ReLU activation, BatchNorm, and Dropout (dropout rate: 0.25), with an output dimension of $64 \times 2$ (for binary classification, e.g., sex). Similarly, the regressor follows the same structure, but the final output layer has a dimension of $64 \times 1$. 
\end{itemize}

Cross-entropy loss was used for classification, while MSE was used for regression. Random search was used for choosing the best hyperparameters from a set of hyperparameters. The Adam optimizer~\citep{kingma2014adam} was used for model training with a learning rate of $1e^{-3}$, running for 200 epochs with a batch size of 64. 

\end{document}